\documentclass[sn-mathphys-num]{sn-jnl}
\usepackage{graphicx}%
\usepackage{multirow}%
\usepackage{xcolor, colortbl}
\usepackage{amsmath,amssymb,amsfonts}%
\usepackage{amsthm}%
\usepackage{mathrsfs}%
\usepackage{tabu}
\usepackage[title]{appendix}%
\usepackage{xcolor}%
\usepackage{textcomp}%
\usepackage{manyfoot}%
\usepackage{booktabs}%
\usepackage{algorithm}%
\usepackage{algorithmicx}%
\usepackage{algpseudocode}%
\usepackage{listings}%
\usepackage{graphicx}
\usepackage{subfigure} 
\usepackage{lineno}
\begin{document}

\title[Article Title]{A Personalized Video-Based Hand Taxonomy: Application for Individuals with Spinal Cord Injury}

\author[1,2,3,4]{\fnm{Mehdy} \sur{Dousty}}\email{Mehdy.Dousty@mail.utoronto.ca}
\author[4,5]{\fnm{David J.} \sur{Fleet}}\email{Fleet@cs.utoronto.ca}
\author*[1,2,3,6]{\fnm{Jos\'e} \sur{Zariffa}}\email{Jose.Zariffa@utoronto.ca}

\affil[1]{\orgdiv{ Institute of Biomedical Engineering}, \orgname{University of Toronto},  \city{Toronto,, \postcode{M5S 3G9}, \state{ON}, \country{Canada}}}

\affil[2]{\orgdiv{Edward S. Rogers Sr. Department of Electrical and
Computer Engineering}, \orgname{University of Toronto, \city{Toronto}, \postcode{M5S 3G9}, \state{ON}, \country{Canada}}}

\affil[3]{\orgdiv{KITE Research Institute},\orgname{University Health Network},\city{Toronto}, \postcode{M5G 2A2},\state{ON},\country{Canada}}

\affil[4]{\orgdiv{Vector Institute}, \city{Toronto}, \postcode{M5G 1M1}, \state{ON}, \country{Canada}}

\affil[5]{\orgdiv{ Department of Computer Science}, \orgname{University of Toronto}, \city{Toronto}, \postcode{M5S 2E4}, \state{ON}, \country{Canada}}

\affil[6]{\orgdiv{Rehabilitation Sciences Institute},\orgname{University of Toronto}, \city{Toronto}, \postcode{M5G 1V7}, \state{ON}, \country{Canada}}

\abstract{ Hand function is critical for our interactions and quality of life. Spinal cord injuries (SCI) can impair hand function, reducing independence. A comprehensive evaluation of function in home and community settings requires a hand grasp taxonomy for individuals with impaired hand function. Developing such a taxonomy is challenging due to unrepresented grasp types in standard taxonomies, uneven data distribution across injury levels, and limited data. This study aims to automatically identify the dominant distinct hand grasps in egocentric video using semantic clustering. Egocentric video recordings collected in the homes of 19 individual with cervical SCI were used to cluster grasping actions with semantic significance. A deep learning model integrating posture and appearance data was employed to create a personalized hand taxonomy. Quantitative analysis reveals a cluster purity of $67.6\% \pm 24.2\%$ with with $18.0\% \pm 21.8\%$ redundancy. Qualitative assessment revealed meaningful clusters in video content. This methodology provides a flexible and effective strategy to analyze hand function in the wild. It offers researchers and clinicians an efficient tool for evaluating hand function, aiding sensitive assessments and tailored intervention plans.}

\maketitle
\section{Introduction}\label{sec1}

The human hand is a vital organ essential for tool use and interaction with the environment. Hand function is crucial for daily independence and thus a top priority for recovery for individuals with cervical spinal cord injuries (SCI) \cite{anderson2004targeting}. Current clinical assessments focus on standardized tests in controlled settings \cite{kalsi2012graded}, but do not always reflect hand function in everyday activities \cite{lang2023improvement}. In clinical settings, individuals are observed grasping objects with specific shapes, weights, and orientations, typically from uncluttered and easily accessible surfaces \cite{cini2019choice}. In contrast, real-life scenarios involve diverse objects with unique features and textures, located in various positions and heights, and used for multiple purposes. As such, clinical observations do not fully describe hand function in home and community environments. Hence, it is necessary to create novel approaches for evaluating hand function in real-life situations \cite{lang2023improvement} to enhance comprehension of its significance in the daily interactions and independence of individuals with SCI.

Egocentric video, captured from first-person wearable cameras, is a valuable tool for studying hand function in natural environments \cite{dousty2023grasp}. This modality is unique among wearable technologies in its ability or provide contextual information for the movements performed. By analyzing these videos, researchers can track the types of grasps \cite{dousty2020tenodesis, dousty2023hand, bandini2022measuring} and their frequency during activities of daily living (ADLs), providing crucial information on the recovery of hand functions \cite{dousty2023grasp}. A recent study \cite{dousty2023grasp} demonstrated a significant link between established clinical assessments and the adoption of particular grasping techniques in domestic settings, validating the importance of analyzing grasping strategies at home. Conversely, certain grasping methods appeared unrelated to clinical scores, indicating that home environments offer additional insights beyond what clinical settings can capture, and that clinical assessments provide only a limited representation of hand function in everyday contexts.

Processing large volumes of video data necessitates automated methods, requiring a clear conceptual framework. Grasp taxonomies can provide valuable guidance in this respect \cite{feix2015grasp}, but the impairment of hand function affects both fine and gross motor control, potentially leading to the development of new grasping strategies not captured by conventional taxonomies. Designing an effective hand taxonomy to capture hand function in neurorehabilitation faces challenges due to altered hand kinematics and difficulties in exerting necessary force post-impairment \cite{ hermsdorfer2003grip}. Traditional methods such as thumb position or virtual finger \cite{feix2015grasp} may not adequately label different hand grasps in the presence of impairment. Creating a new hand taxonomy for individuals with hand impairment is challenging due to considerable variations in function and postures depending on the type and severity of impairment \cite{bensmail2010botulinum}, coupled with the difficulty of collecting large datasets in neurorehabilitation contexts. 

An alternative approach to creating a comprehensive hand taxonomy involves using unsupervised learning techniques to discover reoccurring grasps in video data and group them into homogeneous partitions based on semantic similarity \cite{huang2015we,dousty2020towards}. This strategy offers flexibility to accommodate the varied types of grasp resulting from impairment, tailored to each individual, without relying on predefined taxonomies. However, defining a meaningful similarity or distance function for clustering without true labels presents challenges, as irrelevant aspects in images, such as background information or specific colors, can confuse the clustering process \cite{domingos2012few, aggarwal2001surprising}. This problem is compounded by the 27 degrees of freedom of the human hand, which creates a high-dimensional space within which similar postures must be defined.

 Deep convolutional neural networks \cite{lecun2015deep} have demonstrated outstanding performance in a variety of visual recognition tasks. Such networks can learn to extract features that are generic enough to generalize well for previously unseen data \cite{hu2017learning, guerin2017cnn}. These extracted features can be employed to cluster image data. These methods typically involve concurrent (end-to-end) \cite{ wang2020sa,shiran2021multi} or sequential feature construction and clustering \cite{guerin2021combining, guerin2017cnn, hu2017learning}. Although end-to-end methods exhibit promising results in small datasets such as MNIST and USPS \cite{genevay2019differentiable}, their performance diminishes in larger real-world datasets with cluttered scenes and multiple objects \cite{li2018discriminatively}. In contrast, sequential deep clustering methods, which rely on previously learned knowledge \cite{gong2015web, guerin2017cnn, guerin2021combining}, can outperform end-to-end approaches, particularly when training data are limited \cite{guerin2021combining}.

\begin{figure}[!htb]
    \centering
    \includegraphics[width=\columnwidth]{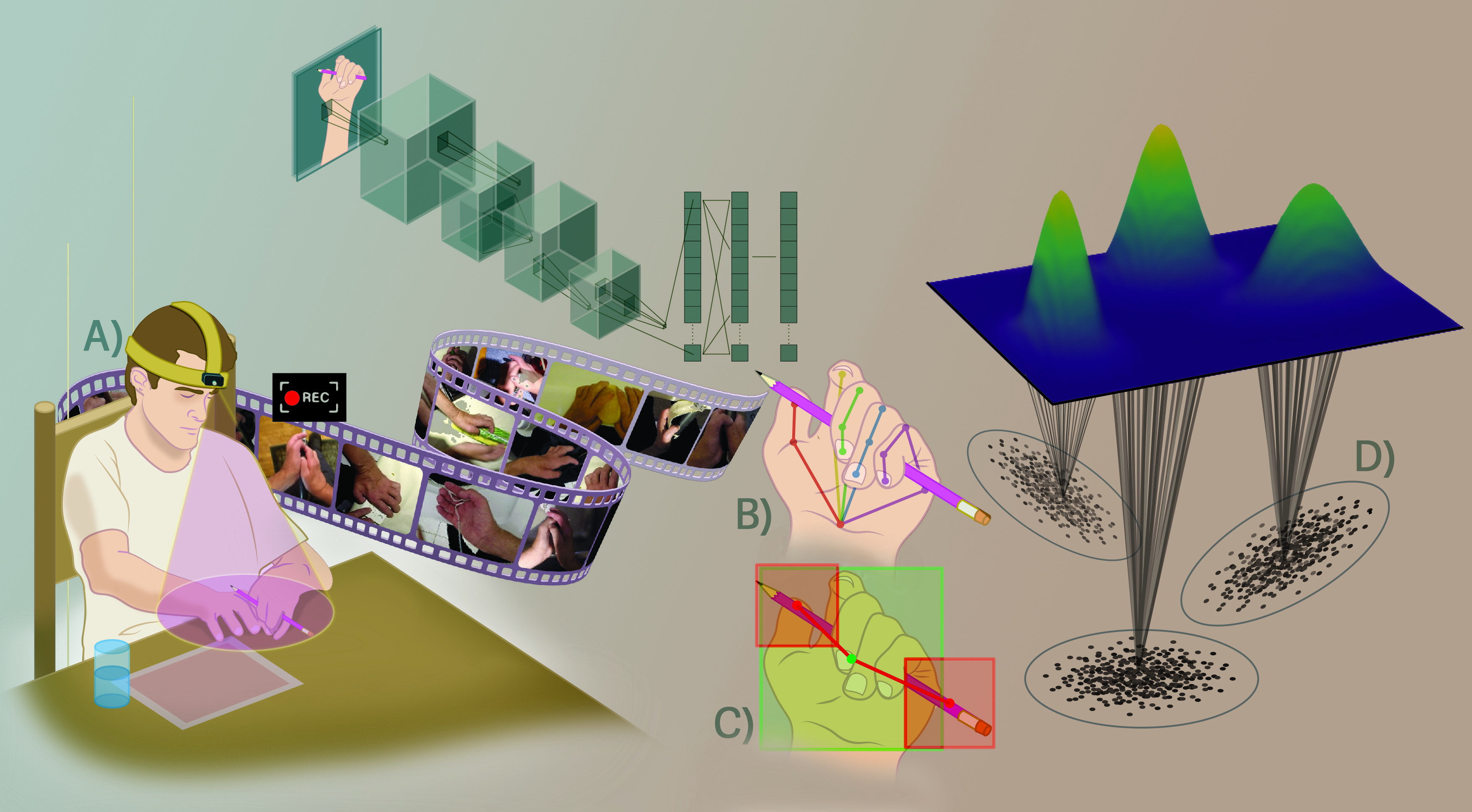}
        \caption{The objective and pipeline of the paper: A) A person records a video with an egocentric camera in a naturalistic home environment. Deep learning methods were employed to extract B) hand poses and C) appearance information encoding hand-object association. D) Clustering algorithms were applied to categorize different grasp types using both hand poses and appearance information.}
      \label{fig:deep}
\end{figure}

The objective of this study, illustrated in Figure \ref{fig:deep}, was to group similar grasps at the individual level in egocentric video recordings, using clustering techniques. These results were then used to condense video data based on predominant grasps post-SCI, marking the first introduction of a clustering approach to develop personalized hand taxonomies. Using annotated data sets, suitable features and clustering techniques were identified to categorize similar grasping actions, with effectiveness demonstrated on unlabeled real-world data. In addition, a novel keyframe selection method was proposed to create a report that can facilitate the analysis of hand function. The findings offer clinicians and researchers, for the first time, a direct window into grasping strategies at home, with applications in personalized treatment planning and sensitive assessment. The key contribution of this work is a methodology to analyze hand use in the wild without restriction to predefined grasp taxonomies.

\section{Results}\label{sec2}
In the context of static hand grasp clustering, two crucial types of information are essential: postural information, and the association between hand postures and interacting objects \cite{dousty2023grasp}. The combination of these types of information can be approached through multiview clustering (MVC), which leverages multiple views from the same image \cite{guerin2021combining}. To ensure success in complex real-world scenarios, thorough empirical examination is necessary, including selecting appropriate deep models, integrating extracted information, and quantifying clustering performance.

\begin{figure}[h!]
    \centering
    \includegraphics[width=\columnwidth]{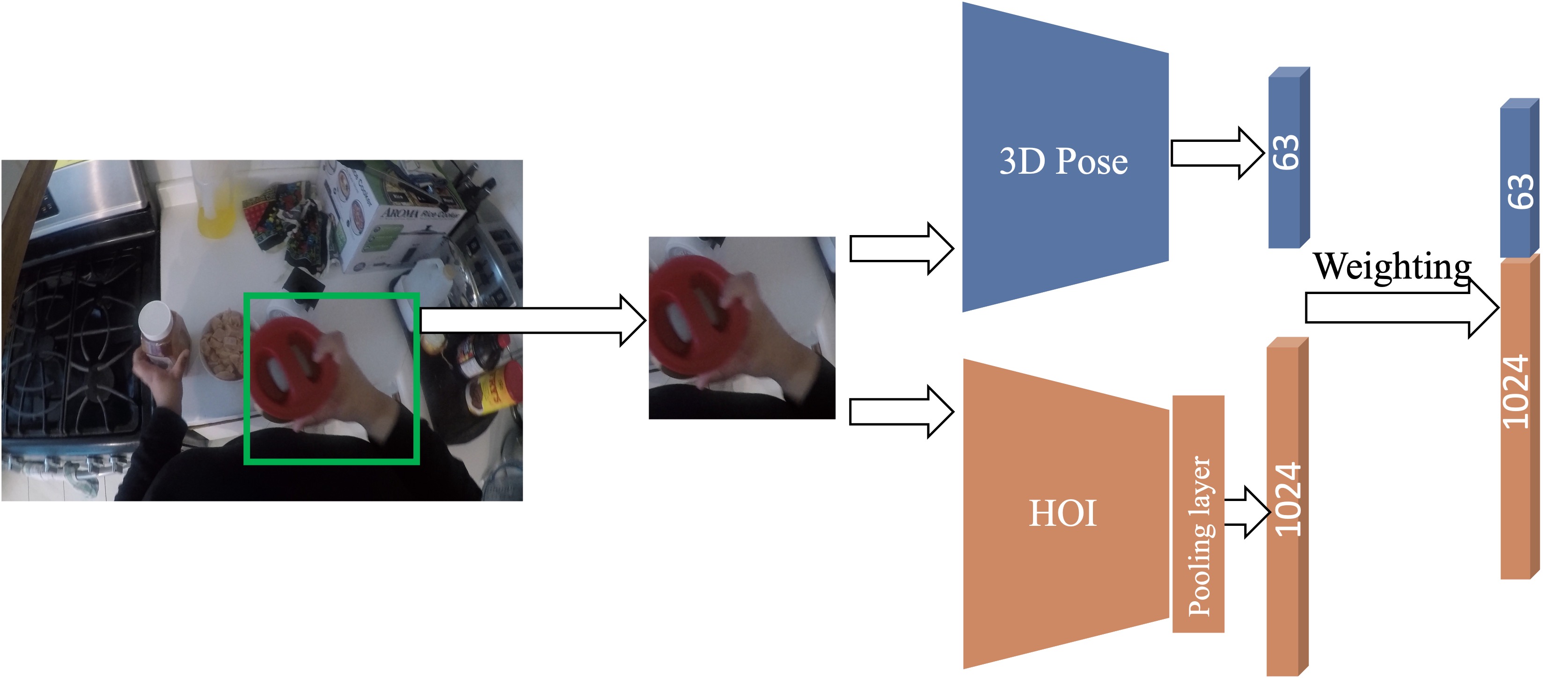}
        \caption{MVC using postural and appearance information }
      \label{fig:MVC}
\end{figure}

Our method uses clustering techniques on combined posture and appearance features. Postural data are derived from 2D and 3D pose estimation algorithms, with 2D joint coordinates and their confidence score estimated using OpenPose \cite{simon2017hand} and 3D joint coordinates estimated using MeshTrans \cite{lin2021end}. Appearance information is extracted from intermediate layers of pre-trained deep models applied to the RGB images. Appearance features are obtained using a Hand-Object Interaction (HOI) network, which is a model to predict hand-object contact \cite{shan2020understanding} after estimating the hand bounding box \cite{visee2020effective}. Normalized postural and appearance data (see Methods) were combined using an intermediate fusion approach to multiview clustering (Figure \ref{fig:MVC}). A hyperparameter was used to control the balance between appearance and postural information, and postural coordinates were weighted with values ranging from 0.001 to 100.0. 

Our investigation involved the following steps, detailed in subsequent sections:

\begin{enumerate}
    \item Methodology optimization on an annotated dataset.
    \item Hand grasp clustering in the wild.
\end{enumerate}

This study used two datasets collected in previous studies: HomeLab-ANS \cite{likitlersuang2019egocentric} and Home-ANS \cite{bandini2022measuring}. HomeLab-ANS includes videos from a head-mounted camera worn by 17 individuals with SCI performing daily tasks in a simulated home environment without specific instructions \cite{likitlersuang2019egocentric} (this dataset is also referred to as "ANS-SCI" in previous work, but we use different nomenclature here for clarity in distinguishing the datasets). Home-ANS comprises videos from a head-mounted camera worn by 21 individuals with SCI as they executed daily tasks in their homes over two weeks (204 ± 42 minutes per participant) \cite{bandini2022measuring}. For the purpose of selecting hyperparameters, a small subset of both data sets was annotated with four primary hand grasps: power, precision, intermediate and nonprehensile, for participants from the Home and HomeLab data sets \cite{dousty2023hand,dousty2023grasp}. For both dataset a uniform sampling technique was employed to mitigate data imbalance, collecting a fixed number of frames per task regardless of duration. 

\subsection{Methodology optimization on an annotated dataset}
Various clustering algorithms (Method section) were evaluated, including partitional clustering with the k-means algorithm, Gaussian mixture models (GMM), and spectral clustering methods. Cluster performance was evaluated using internal and external evaluation indices (IEI and EEI) \cite{jain1999data}, including the Silhouette coefficient, maximum match (MM), Fowlkes-Mallows (FLK) and normalized mutual information (NMI) \cite{saxena2017review}. The search space for clustering algorithms is vast. A heuristic approach was therefore used to identify the clustering method with the best performance by examining results on the HomeLab dataset using 2D postural information features(\ref{tab:clusterhyper}). Hyperparameter optimization showed that K-Means with an L-1/2 distance function and GMM with a tied covariance shape had the highest performance. Next, we compared those clustering algorithms using several feature spaces in both annotated datasets (\ref{tab:clusterhyper2},\ref{tab:Bclusterhyper1}). GMM was chosen for further analysis due to its lower computational demands and good performance.

Next, our aim was to identify the feature space with the highest performance in terms of IEI and EEI. This involved using an annotated dataset that described the four main grasp types. These four grasp annotations may encompass different grasp configurations; for instance, a power grasp might consist of a fixed hook or a large diameter grasp \cite{feix2015grasp}. These varied configurations can result in sparsity within the feature space for a specific grasp type. We employed a cluster analysis approach to address this issue by varying the number of clusters up to 10. We evaluated the performance of different feature spaces by calculating the Silhouette score and selected the one with highest score. 

Our observations suggest that, on average, including confidence scores in 2D pose estimation was advantageous, and the sampling method generally improved the performance of IEI (\ref{tab:clusterhyper21}). We also demonstrated that 3D pose features yield higher performance compared to 2D pose features (\ref{tab:clusterhyper2}). We found that global average pooling achieved higher average IEI and EEI scores (\ref{tab:Bclusterhyper1}).

Our feature analysis also included an attempt to identify features that remained consistent despite input augmentation. We employed self-supervised learning using the BYOL algorithm \cite{grill2020bootstrap} to minimize variations in different views of an image through temporal and spatial augmentation (\ref{fig:method}). However, we were unable to observe consistent improvements across various clustering metrics (\ref{tab:Bclusterhyper2}). A detailed description of this method is available in the supplementary section of our work (\ref{secA2}). 

\begin{figure}[!htb]
    \begin{center}
    \centering
    \includegraphics[scale=0.080]{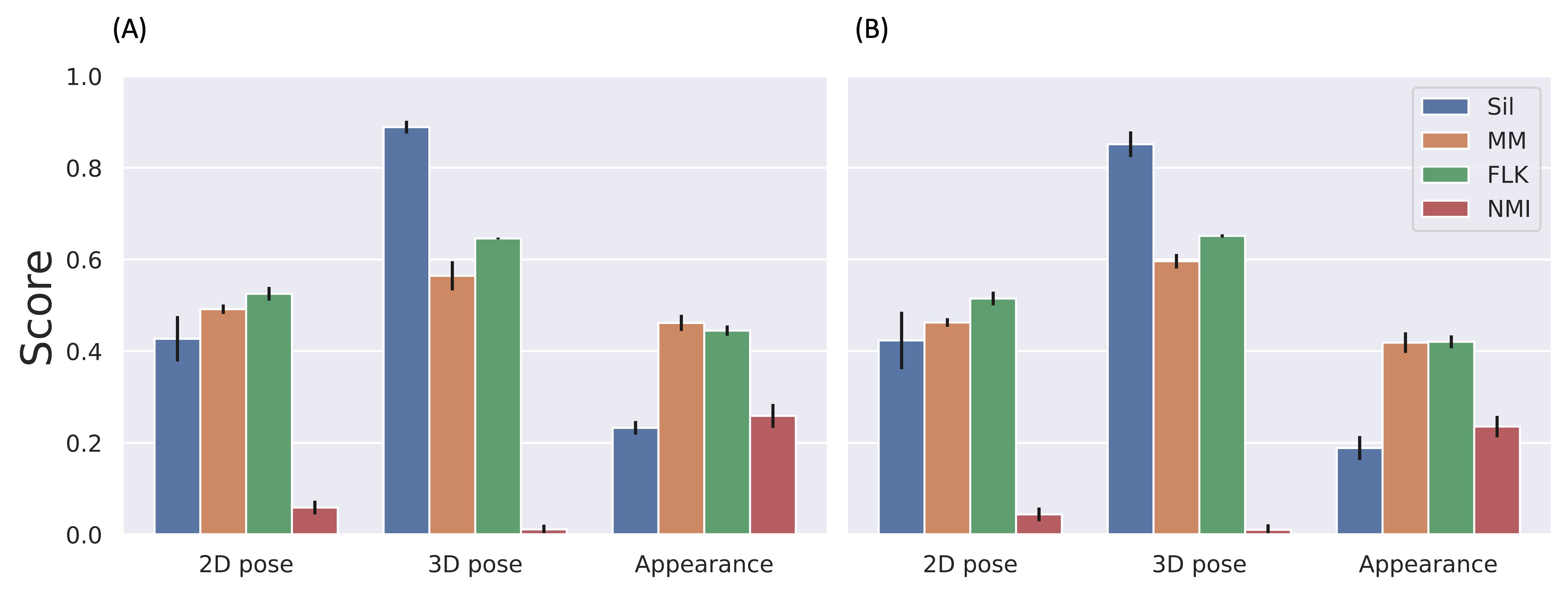}
    \end{center}
    \caption{EEI and IEI for 2D pose, 3D pose, and appearance using GMM algorithm with tied covariance for A) HomeLab (n = 17 participants) and B) Home dataset (n = 19 participants), considering mean and standard errors.}
    \label{fig:pose_app}
\end{figure}

The results shown in Figure \ref{fig:pose_app} illustrate the results of our analysis, including the 2D pose, the 3D pose, and the appearance information using a GMM approach with a tied covariance shape. They show that, as a general trend, postural information tends to exhibit elevated Silhouette scores, MM values, and FLK metrics. On the contrary, appearance information tends to showcase higher NMI values. This may suggest that appearance information may provide different insights compared to postural information. 

\begin{figure}[!h]
    \centering
    \includegraphics[width=\textwidth]{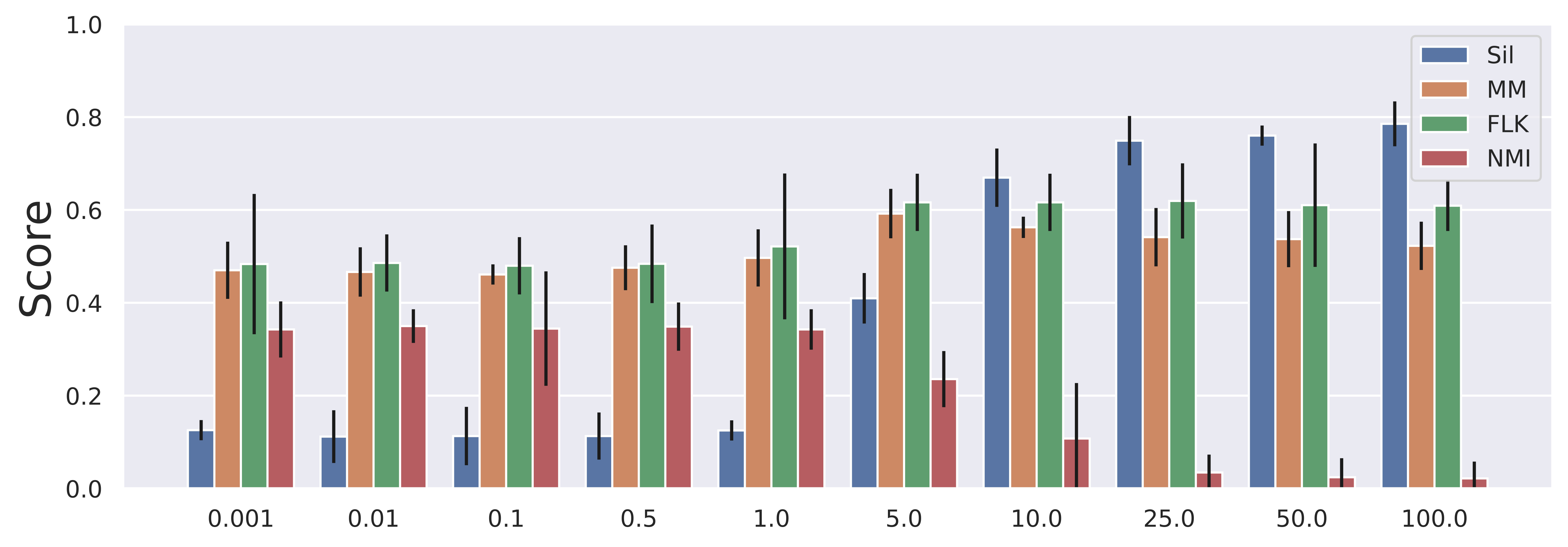}
    \caption{EEI and IEI using different weighting for postural information using GMM algorithm for Home dataset (n = 19 participants), considering mean and standard errors.}
    \label{fig:wpose}
\end{figure}

We combined 3D postural information with appearance data using various weighting coefficients detailed in Figure \ref{fig:wpose}. Postural weighting notably influenced the Silhouette score and NMI more than other metrics. An inverse correlation between the Silhouette score and NMI was observed, aligning with earlier observations of NMI's sensitivity to cluster count. A high Silhouette score can be misleading, as it may force the entire dataset into a single large cluster, leading to a high Silhouette score but a low NMI score and a collapse in the feature space. To mitigate this, we opted for a weighting of 5.0, balancing NMI and the Silhouette score, yielding reasonably acceptable performance. Similar behavior was observed in the HomeLab dataset. The findings suggest that combining postural and appearance information provides a more favourable trade-off across the different performance metrics examined.

\begin{figure}[!h]
    \centering
    \includegraphics[width=\textwidth]{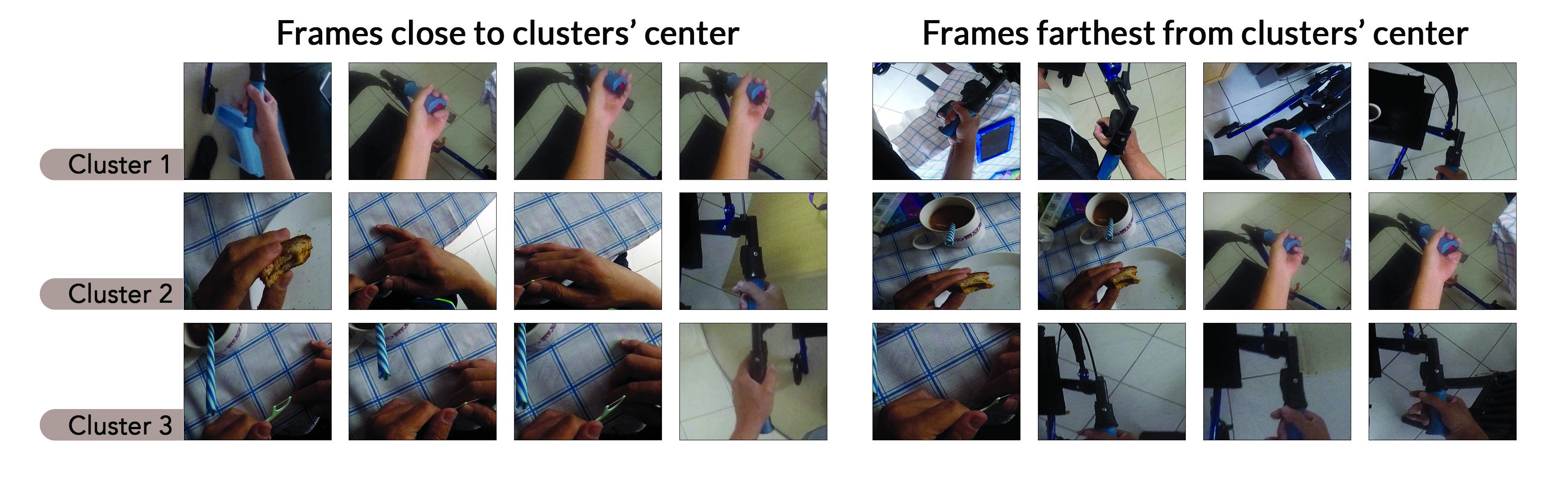}
    \caption{Qualitative example of grasp clustering from participant 10 in the labelled subset of the Home dataset.}
    \label{fig:frameannot}
\end{figure}

We also present qualitative results for participant 10. Clustering results, shown in Figure \ref{fig:frameannot}, displays the frames found close and farthest to cluster centers for one participant for the three estimated clusters. The qualitative findings indicate that the model exhibits significant potential in categorizing various grasping patterns. Cluster 1 primarily contains frames showcasing powerful actions, while cluster 2 predominantly emphasizes precision, and cluster 3 focuses mainly on intermediate grasping. As we move farther away from the centers of these clusters, the model's accuracy decreases, leading to more errors. The dataset for this participant had limited variability in tasks and locations, but despite similarities in backgrounds and objects, the clustering process was still able to identify distinct grasps.

\subsection{ Hand Grasp Clustering in the Wild}
The annotations used above to evaluate clustering methods provide a broad classification of hand function but do not capture the full spectrum of grasping strategies present in the dataset. The next stage of our analysis therefore focused on the full unlabelled dataset. We applied a hand-object interaction model \cite{shan2020understanding} to distinguish between interaction and non-interaction video segments. We smoothed the binary frame-level outputs of hand-object interaction detection with a median filter of size 17 and divided each interaction into 16 intervals, randomly selecting frames within each interval. This approach helped create a balanced dataset with an average of 994.75 +/- 112.3 interactions per participant, including approximately 19,807 +/- 1,203 frames with right-hand interactions and 19,983 +/- 1,512 frames with left-hand interactions.

We estimated the number of clusters in the dataset using the elbow method with the Bayesian Information Criterion excluding clusters with fewer than 10 frames. Clustering performance was measured using purity and redundancy \cite{arinik2021characterizing}. To simplify the evaluation process, we sampled 30 frames from each predicted cluster, selecting one random frame for every group of N/30 frames (where N represents the total number of cluster members) after sorting by distance to the cluster centre. We assessed purity by manually examining the 30 sampled frames for similar hand configurations and contact points. Redundancy was quantified by identifying clusters with more than 50\% of frames shared with similar grasps, dividing the number of redundant clusters by the total number of clusters.

\begin{figure}[!h]
    \centering
    \includegraphics[width=\textwidth]{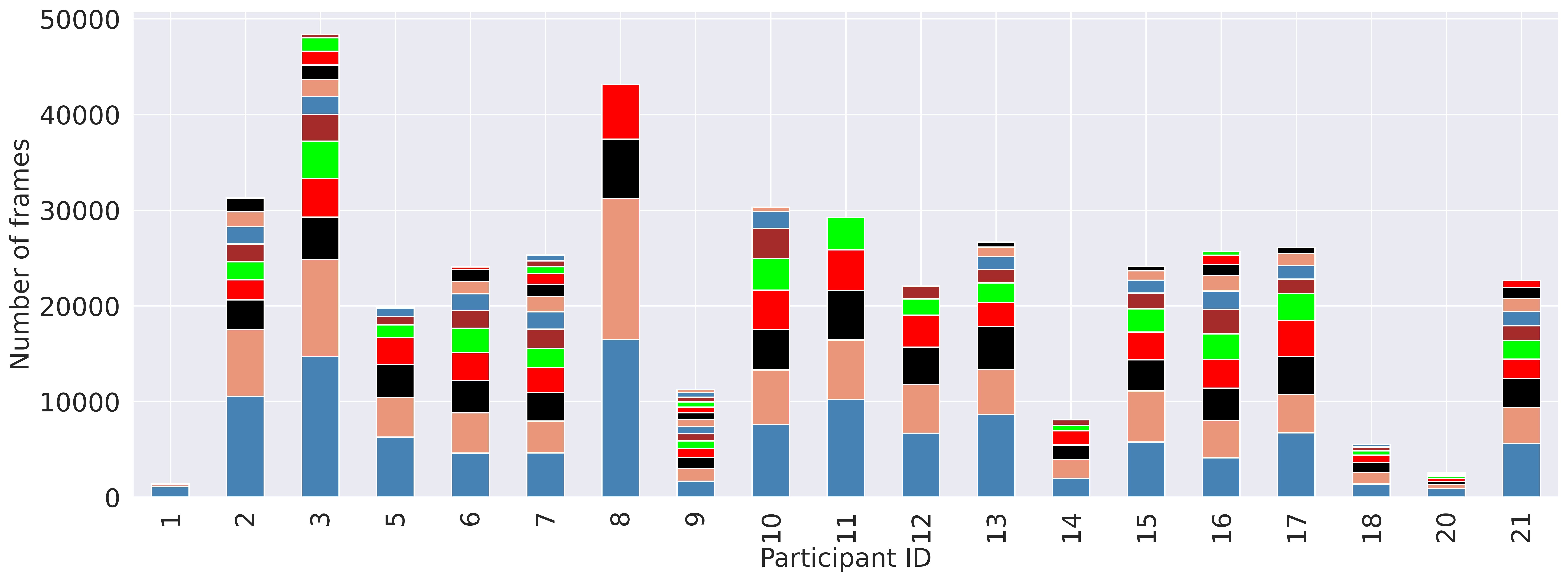}
    \includegraphics[width=\textwidth]{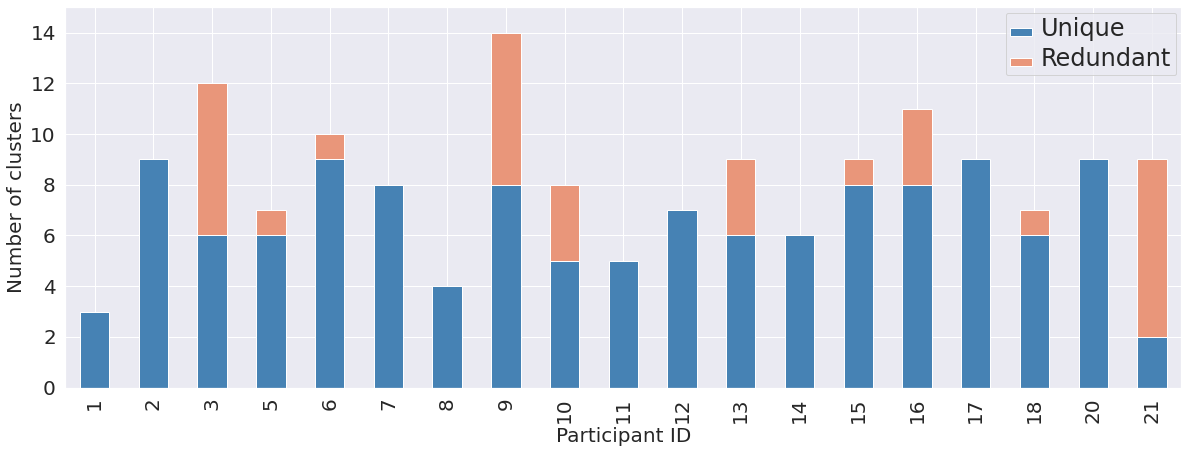}
    \includegraphics[width=\textwidth]{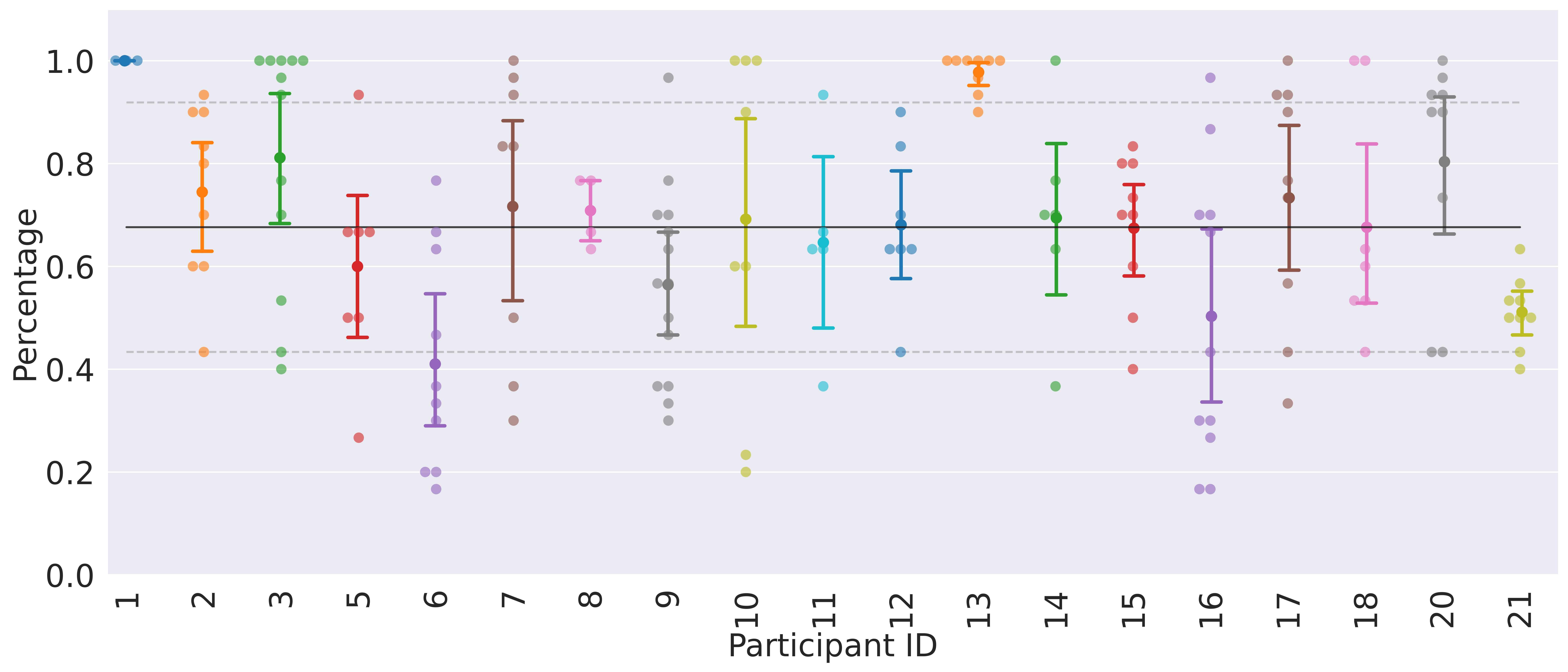}
    \caption{Clustering performance in the wild: A) Number of frames per cluster. Different colors represent distinct clusters. B) Count of unique and redundant clusters for each participant. C) Purity per cluster for each participant.}
    \label{fig:ResF}
\end{figure}

Based on our findings from the annotated datasets, we applied the GMM algorithm to 19 participants in the unlabelled Home dataset, excluding participants 4 and 19. Participant 4 did not complete the study, and Participant 19's hand-object interaction data was insufficient for analysis. The outcomes of our quantitative analysis are presented in Figure \ref{fig:ResF}. On average, $8.2 \pm 2.6$ clusters per participant, with $18.0\% \pm 21.8\%$ redundancy and $67.6\% \pm 24.2\%$ purity exist in the dataset.

\begin{figure}[]
    \includegraphics[width=\textwidth]{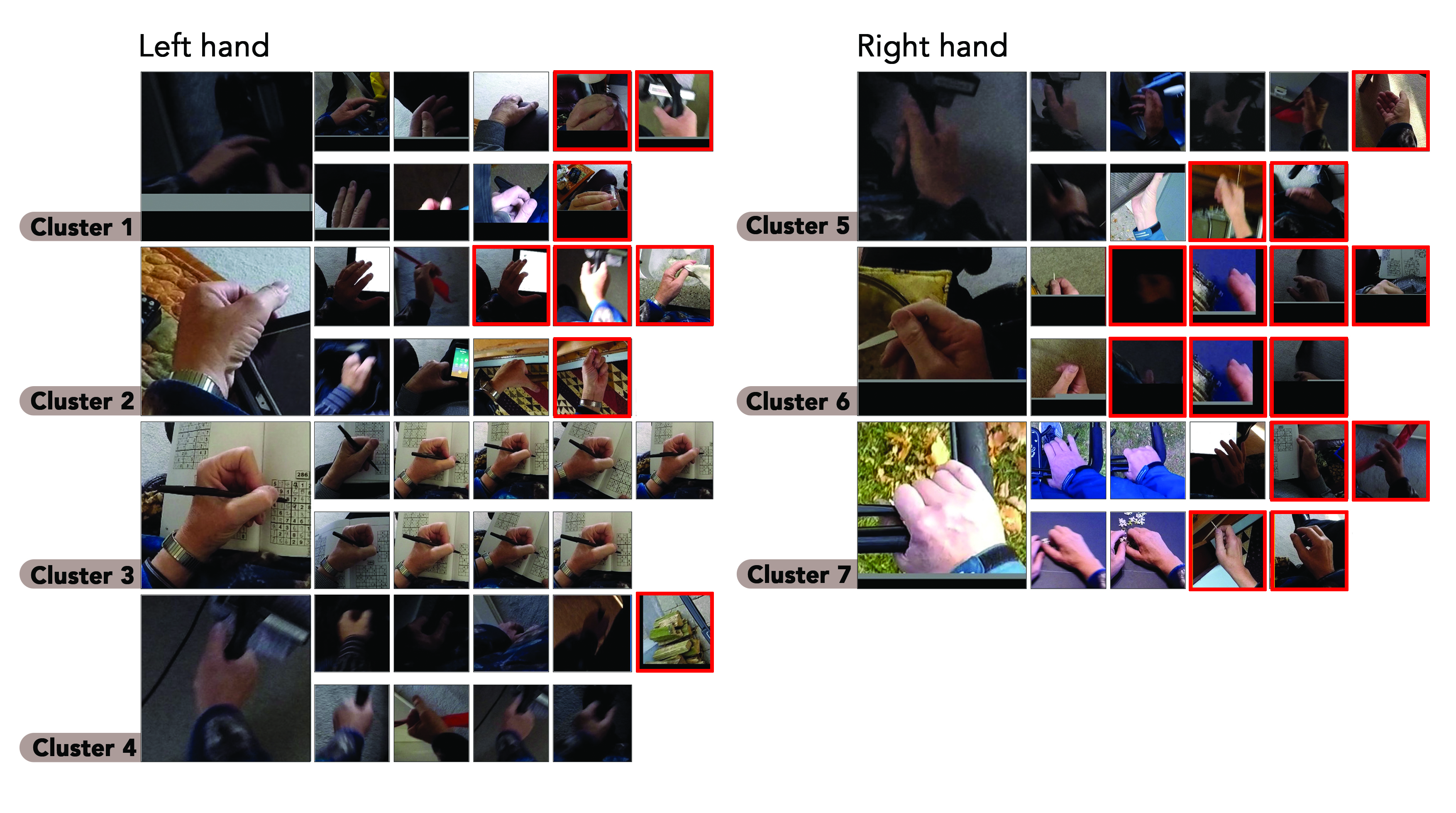}
    \hrule
    \includegraphics[width=\textwidth]{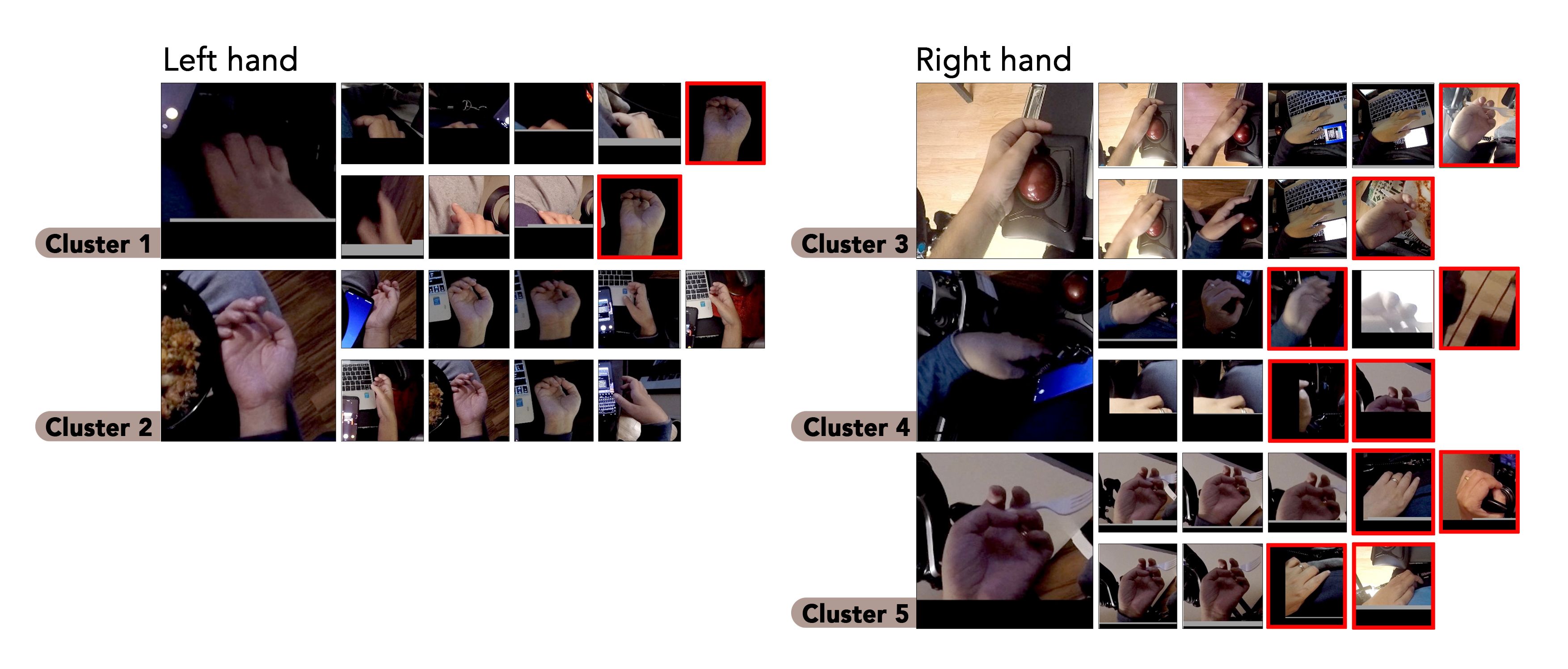}
    \hrule
    \includegraphics[width=\textwidth]{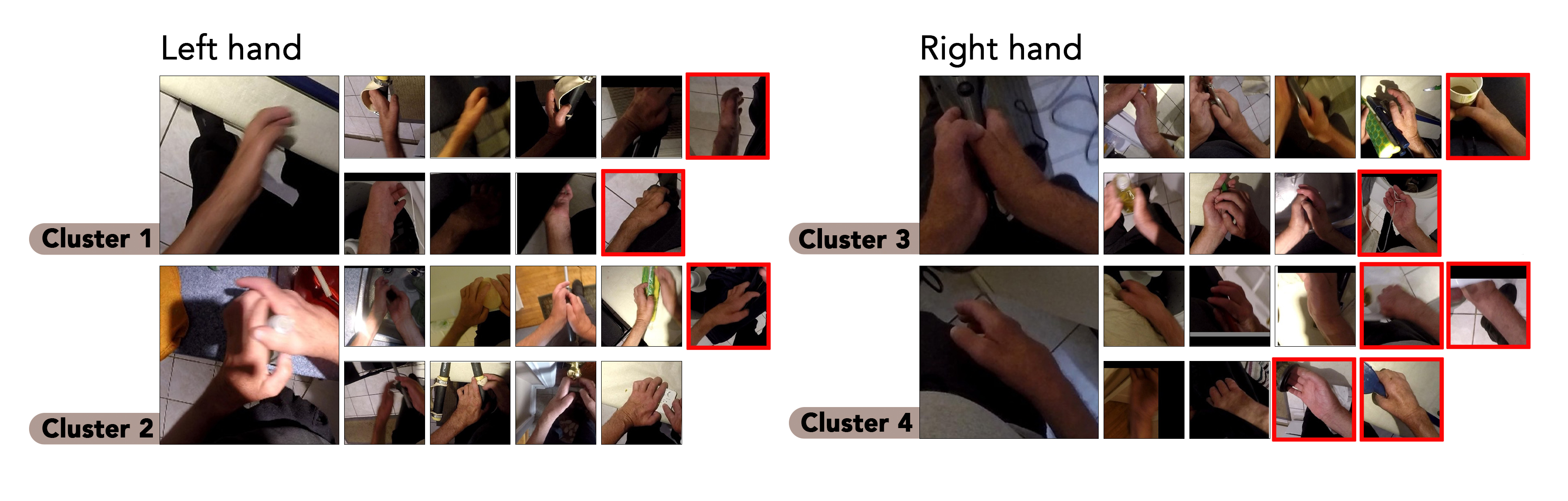} 
    \caption{ Seven clusters, five clusters, and four clusters were estimated for Participant 12, 11, and 8, respectively, with the central point of each cluster displayed within the large frame. The grasps that do not correspond to the same grasp type as the cluster center are represented using red bounding boxes.}
    \label{fig:R1}
\end{figure}

The grasp categories identified by the clustering algorithm are not given specific names. To create a report that clinicians or researchers can quickly analyze, we provide a few examples per cluster that best represented each cluster. The image selection process considered both the distance to the cluster center and the image quality. Specifically, the images were assessed and sorted based on their likelihood of being associated with a specific cluster. 50\% of the frames with the highest likelihood belonging to the cluster were chosen, and these selected images were then divided into ten intervals. From each interval, a random image with high image quality was chosen, ensuring that images from the same interaction were eliminated in the other intervals. Image quality was evaluated by analyzing the variance of the image Laplacian. Figure \ref{fig:R1} present a visual representation of each cluster's keyframes. These visuals can serve as a valuable report, showcasing an individual with SCI's predominant hand postures in their everyday context.

\section{Discussion}\label{sec3}
Grasping strategies are dependent on factors relating to the individual, the task, and the environment. Methods to analyze hand use in a real-world context are therefore of significant interest. Traditional hand taxonomy construction relies on a comprehensive categorization of grasp types, which may overlook new grasps due to impairment. Designing new taxonomies for a neurorehabiltiation context (or similarly for other situations with specialized grasps) is challenging due to post-injury variability in function. We have demonstrated that egocentric video combined with unsupervised learning can be used to group similar grasps and obtain a personalized summary of grasping strategies. 

The proposed method offers a significant advantage over grasp classification using hand taxonomies developed for non-impaired individuals, by adapting to each individual's impairment characteristics. This personalized hand taxonomy enables efficient analysis of hand function through long-term video capture of real ADLs at home. Clustering results can serve as a summarization technique for clinicians or researchers, providing insights into hand impairments, enabling sensitive assessments, and aiding in tailored rehabilitation programs by designing exercises that target specific grasps and support the restoration of maximum hand function. Having a better grasp can result in increasing independence \cite{dousty2023grasp} and eventually quality of life. This novel capability could potentially also provide researchers with enhanced opportunities to investigate hand function. For instance, researchers might explore the connection between the quantity and compactness of clusters and how they relate to the recovery profile.

This study used a sequential deep clustering method, incorporating postural and appearance information to cluster images of similar grasps. However, the performance of this approach, particularly when using only appearance information, is influenced by several factors. These factors include the architecture of the CNN backbone, the specific layers from which features are extracted, and the dataset used for training the model \cite{raghu2021vision, cohen2020separability,somepalli2022can, guerin2021combining, sharif2014cnn}. The extracted features' accuracy relies on the similarity between the data distribution used for feature extraction and the data distribution used for training the model. Changes in data distribution can lead to a shift in the feature distribution, potentially reducing the accuracy of the extracted features \cite{koh2021wilds, guerin2021combining, guerin2021combining}. This is because most of the training and testing data lie close to the decision boundary of the model, and changes in the data distribution can affect the shape of this boundary \cite{mickisch2020understanding}. For example, a CNN model trained on the ImageNet dataset\cite{deng2009imagenet} will perform better on natural object clustering tasks similar to the ImageNet dataset than on scene recognition tasks that are not \cite{guerin2021combining}. To address this, we employed the hand-object interaction model trained on a large dataset to extract relevant features. One possible approach to enhance the quality of these extracted features is to retrain and fine-tune the model to adapt it to the new data distribution. We found that using global average pooling instead of max pooling led to better performance, contradicting previous research suggesting max pooling enhances discriminability \cite{boureau2010theoretical}. Max pooling is more noise-sensitive, resulting in sparser representations and lower Silhouette scores in clusters. In contrast, average pooling is less noise-sensitive, creating more compact and confident latent spaces and contributing to higher Silhouette scores. The choice of other pooling layers \cite{lee2016generalizing} in the sequential clustering method is a promising area for future research.

Our study demonstrated that leveraging coordinates obtained through 3D pose estimation and features derived from pre-trained HOI models enhances overall performance. To further elevate performance, several future avenues can be explored. For instance, adopting future hand pose estimation techniques that are more resilient to occlusion could yield better results. Training a new HOI model that explicitly extracts contact points could also improve appearance-based features. Another avenue worth investigating is a multi-learner system to combine clustering outcomes generated exclusively from postural and appearance data as an ensemble method. These strategies hold potential for advancing the accuracy and robustness of the clustering process.

Ultimately, the tied GMM shape was chosen for further analysis in the paper. One important drawback of the proposed method was the reliance on a heuristic approach to select the clustering parameters, which may not be optimal for all feature spaces. In future studies, it will be necessary to select hyperparameters for each specific feature space instead of using a fixed 2D pose estimation algorithm. Investigating the influence of clustering hyperparameters on the results will be a key focus of future research. In this work, we also excluded density-based clustering algorithms because determining the minimum number of samples required for clustering introduces an additional optimization parameter, thereby increasing the overall complexity of the process.

Clustering hand grasps in a natural, uncontrolled setting poses several challenges. Firstly, it uses all the data collected during interactions between hands and objects, resulting in a densely populated feature space. This dataset significantly differs from the carefully curated dataset where distinct grasps are explicitly labelled. In the uncontrolled, natural dataset, multiple types of grasps coexist, transitioning between stable and transitional states. Additionally, individuals with SCI often have limited motor control abilities, causing their features to exhibit similar characteristics, thereby complicating the clustering task. Another complication arises from noise introduced through errors in the hand-object interaction and hand detection processes, which can cause features to deviate from the norm in many frames, impeding the clustering process. Having clusters of various sizes is another issue in identifying distinct clusters. Two factors contribute to this issue: 1) having different task durations for different grasp types, and 2) reliance on some grasps more than others. We have attempted to solve the first of these considerations using sampling techniques; however, the latter remains an open question. Despite these complications, we have quantitatively and qualitatively demonstrated that our method is able to identify different numbers of clusters with dominant hand grasps in the right and left hand in multiple hours of video. The model's performance was assessed by choosing samples based on their likelihood to belong to a cluster. We anticipate that increasing the amount of sample data may have a minimal effect on the ultimate results.

The primary reason for choosing SCI as our focus in this study is the diversity of ways in which individuals with SCI use their hands depending on their injury level. This variability creates a rich dataset that our method has shown promise in analyzing. We are confident in its applicability to other conditions affecting hand function, like stroke or Parkinson's disease. Furthermore, this method isn't limited to individuals with hand impairments but can more generally offer a comprehensive data-driven understanding of hand function. This method can find application in different disciplines including ergonomics, sports science and robotics. In conclusion, by enabling a detailed and flexible description of hand use in the wild, this method has the potential to revolutionize our understanding of hand function and greatly improve interventions for individuals with hand impairments. 

\section{Online Methods}\label{sec3}
\subsection{Annotated dataset}
\begin{figure}[!h]
    \centering
    \includegraphics[width=\textwidth]{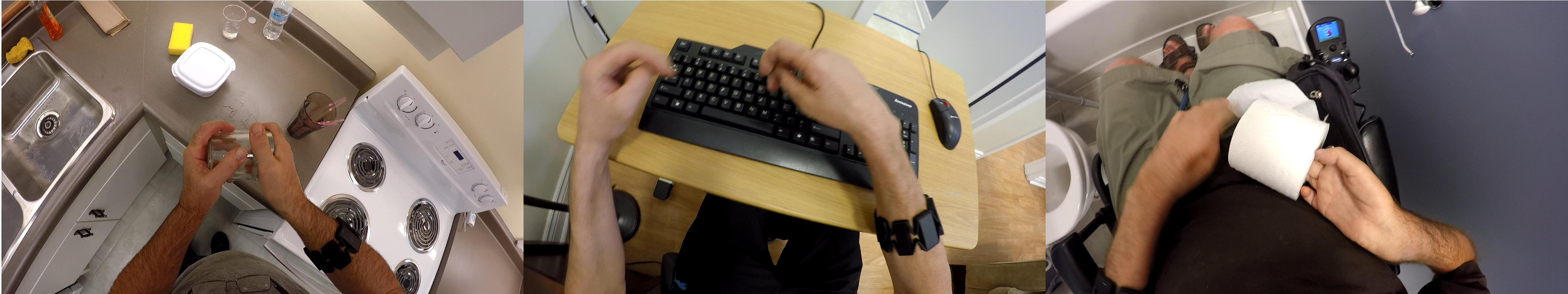}
    \includegraphics[width=\textwidth]{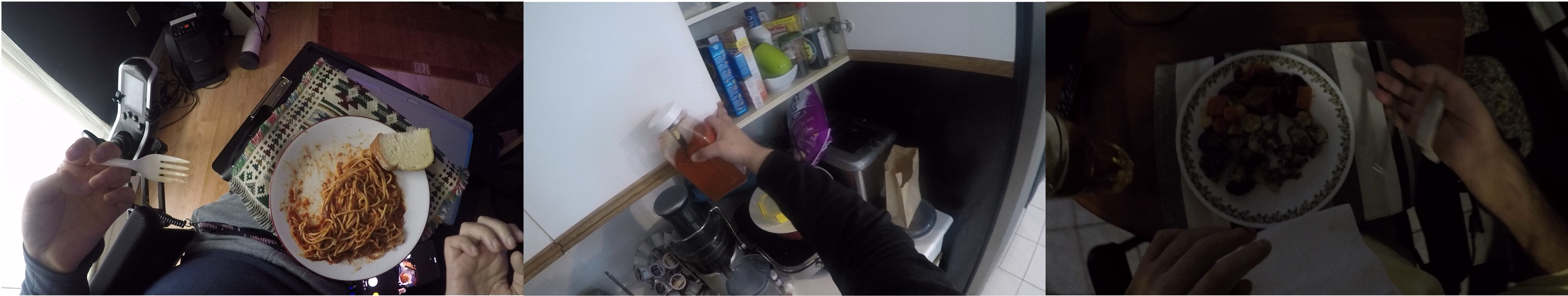}
    \caption{Examples from the dataset of individuals with SCI performing ADLs. The top row shows three examples from the HomeLab-ANS dataset and the bottom row shows three examples from the Home-ANS dataset.}
    \label{fig:exm}
\end{figure}

 Details of the participants and data collection protocols can be found in \cite{likit} for the HomeLab dataset, and in \cite{bandini2022measuring} for the Home dataset. We annotated the datasets using four classes for each participant from the Home and HomeLab dataset: power, precision, intermediate and non-prehensile. In a power grasp, the arm drives the movement of the object, and the perceived force exerted between the hand and the object is perpendicular to the palm. A precision grasp involves the hand moving intrinsically without the movement of the arm, with pressure applied between the distal phalanges. An intermediate grasp is a combination of power and precision grasps in nearly equal proportion, with pressure applied by the intermediate phalanx. In a non-prehensile grasp, there is no net force between the hand and the object, but the net force between the hand and the object is not zero. Examples of non-prehensile grasps include pushing, sliding, rolling an object, and carrying items on a tray.\cite{dousty2023grasp, dousty2023hand}. Table \ref{tab:n_frames} summarizes the resulting frame counts for each dataset and grasp category. Some examples from the dataset are shown in Figure \ref{fig:exm}. The annotation inter-rater reliability metric for the Home dataset, as measured by Cohen’s Kappa score, was 0.81, indicating a strong level of agreement between annotations.\cite{dousty2023grasp}.

\begin{table}[!h]
\centering
\caption{Manually annotated grasp distribution}
\begin{tabular}{
>{\columncolor[HTML]{FFFFFF}}l |
>{\columncolor[HTML]{FFFFFF}}c 
>{\columncolor[HTML]{FFFFFF}}c 
>{\columncolor[HTML]{FFFFFF}}c 
>{\columncolor[HTML]{FFFFFF}}c }
Data/Frames & Power & Precision & Intermediate & Non-Prehensile \\ \hline
HomeLab \cite{likitlersuang2019egocentric}      & 94,987  & 64,201    & 44,717       & 36,062         \\
Home \cite{bandini2022measuring}         & 281,473 & 101,284   & 103,030      & 104,842    \\ \hline    
\end{tabular}
\label{tab:n_frames}
\end{table}

\subsection{Feature extraction}
For the extraction of postural features, we make the clustering process invariant to postural hand translation and rotation by registering the frames such that the wrist is at the origin and the index finger metacarpophalangeal segment is aligned with the y-axis. We used OpenPose, which predicts 2D joint coordinates from images along with their confidence scores \cite{simon2017hand}. The confidence score is useful as it can indicate self- and object occlusion and reveal the contact location. We used these 2D joint coordinates, with and without confidence scores, to construct the feature space for clustering. Next, we normalized the features to have a mean of 0 and a standard deviation of 1 to ensure equal weight in clustering. However, feature normalization may not be optimal \cite{de2016survey} as some fingers may hold greater importance in hand movement classification. For example, the thumb's ability to move in and out is crucial for non-impaired hand taxonomy, whereas the fifth digit may not play a significant role. We evaluated the process of assigning a weight of 5.0 to the thumb joint coordinates. The feature vectors derived from the postural information have a dimensionality of 42 for the 2D postural data without confidence scores, or 63 for 2D postural data. The specific results for both datasets are described in Table \ref{tab:clusterhyper21}. For 3D pose estimation algorithm, only the joint coordinates were used. 

For appearance features, we constructed two feature vectors of dimension 1024 each, using global average pooling and maximum pooling on the Faster RCNN feature map from the HOI model. The features were then normalized to have a mean of 0 and a standard deviation of 1 to ensure equal weight in clustering.

\subsection{Clustering methods}
For partitional clustering, the k-means algorithm was employed \cite{ahmed2020k}, with the k-means++ initialization method \cite{arthur2006k}, a maximum of 300 iterations and a relative tolerance of 1e-4. We also used the Gaussian mixture model (GMM) \cite{fraley1998algorithms} with different covariance shapes examined. Finally, we applied spectral clustering with a different number of neighbours as a graph-based approach. For GMM clustering, we used 100 initializations and allowed a maximum of 300 steps for convergence. Additionally, we adjusted the regularization covariance for each individual if the default value of 1e-6 failed to generate positive covariance matrices. Table \ref{tab:hyperclustering} summarizes the hyperparameters.

\begin{table}[!htb]
\caption{Hyperparameters for clustering algorithm}
\centering
\begin{tabular}{l|l|l}
\rowcolor[HTML]{FFFFFF} 
         & Hyperparameter    & Values                          \\ \hline
\rowcolor[HTML]{EFEFEF} 
K-Means  & Distance function & L 2, L 1, L 1/2, L 1/10         \\
\rowcolor[HTML]{FFFFFF} 
GMM      & Covariance shape  & Full, Spherical, Diagonal, Tied \\
\rowcolor[HTML]{EFEFEF} 
Spectral & Distance function & 20, 40, 60 80                   \\ \hline
\end{tabular}
\label{tab:hyperclustering}
\end{table}

\section{Declarations}
\subsection{Funding}
The work was supported by the Natural Sciences and Engineering Research Council of Canada (NSERC) under Grant RGPIN-2014-05498, the Praxis Spinal Cord Institute under Grant G2015–30, the Ontario Early Researcher award under Grant ER16–12-013, the Craig H. Neilsen Foundation under grant 542675, the Healthcare Robotics (HeRo) NSERC CREATE Graduate Training Program, and the NVIDIA Corporation through the NVIDIA Academic Hardware Grant Program.

\subsection{Ethics approval and consent to participate}
The participants gave written informed consent before taking part in the study. The study was approved by the Research Ethics Board of the institution (Research Ethics Board, University Health Network: 15–8830 and 18-5225.

\subsection{Data availability}
The dataset, comprising de-identified patient data, will be shared exclusively for academic purposes upon a reasonable request and completion of a data sharing agreement.

\subsection{Code availability}
The code will be shared upon request.

\subsection{Author contributions}\label{sec5}
MD, DF, and JZ designed the research. MD conducted data analysis, interpreted the findings, and drafted the manuscript. DF and JZ supervised the project's progress, provided feedback, interpreted the findings, and reviewed the manuscript. 

\subsection{Competing interests}\label{sec6}
The authors declare that the study was conducted without any commercial or financial ties that could be interpreted as a potential conflict of interest.

\begin{appendices}
\section{Hyperparameter tuning}\label{secA1}
\subsection{Cluster hyperparameter tuning}
After conducting the hyperparameter optimization in the HomeLab dataset using 2D postural information, we found that the K-Means algorithm using an L-1/2 distance function performed better than the other methods we studied (Table \ref{tab:clusterhyper}). Among the covariance shapes we explored, the GMM method with a tied covariance shape showed the best overall performance. We also observed that spectral clustering remained effective even when the number of selected neighbours varied. We chose to prioritize these GMM and K-Means methods for further analysis, as they consistently demonstrated superior performance.

\begin{sidewaystable}[]
\caption{Clustering hyperparameter tuning}
\begin{tabular}{lllll}
\rowcolor[HTML]{FFFFFF} 
                                                                        & \multicolumn{4}{c}{\cellcolor[HTML]{FFFFFF}\textbf{KMeans}}                                                                                                                                                                                                                                                                   \\ 
\rowcolor[HTML]{EFEFEF} 
\multicolumn{1}{l|}{\cellcolor[HTML]{FFFFFF}}                           & \multicolumn{1}{l|}{\cellcolor[HTML]{EFEFEF}L2}                               & \multicolumn{1}{l|}{\cellcolor[HTML]{EFEFEF}L1}                               & \multicolumn{1}{l|}{\cellcolor[HTML]{EFEFEF}L1/2}                             & \multicolumn{1}{l|}{\cellcolor[HTML]{EFEFEF}L1/10}                            \\ 
\rowcolor[HTML]{FFFFFF} 
\multicolumn{1}{l|}{\cellcolor[HTML]{FFFFFF}{\color[HTML]{000000} Sil}} & \multicolumn{1}{l|}{\cellcolor[HTML]{FFFFFF}{\color[HTML]{000000} 0.40±0.27}} & \multicolumn{1}{l|}{\cellcolor[HTML]{FFFFFF}{\color[HTML]{000000} 0.54±0.32}} & \multicolumn{1}{l|}{\cellcolor[HTML]{FFFFFF}{\color[HTML]{000000} 0.56±0.29}} & \multicolumn{1}{l|}{\cellcolor[HTML]{FFFFFF}{\color[HTML]{000000} 0.52±0.33}} \\ \hline
\rowcolor[HTML]{EFEFEF} 
\multicolumn{1}{l|}{\cellcolor[HTML]{EFEFEF}{\color[HTML]{000000} MM}}  & \multicolumn{1}{l|}{\cellcolor[HTML]{EFEFEF}{\color[HTML]{000000} 0.46±0.08}} & \multicolumn{1}{l|}{\cellcolor[HTML]{EFEFEF}{\color[HTML]{000000} 0.47±0.08}} & \multicolumn{1}{l|}{\cellcolor[HTML]{EFEFEF}{\color[HTML]{000000} 0.47±0.11}} & \multicolumn{1}{l|}{\cellcolor[HTML]{EFEFEF}{\color[HTML]{000000} 0.46±0.11}} \\ \hline
\rowcolor[HTML]{FFFFFF} 
\multicolumn{1}{l|}{\cellcolor[HTML]{FFFFFF}{\color[HTML]{000000} FLK}} & \multicolumn{1}{l|}{\cellcolor[HTML]{FFFFFF}{\color[HTML]{000000} 0.46±0.10}} & \multicolumn{1}{l|}{\cellcolor[HTML]{FFFFFF}{\color[HTML]{000000} 0.50±0.10}} & \multicolumn{1}{l|}{\cellcolor[HTML]{FFFFFF}{\color[HTML]{000000} 0.53±0.10}} & \multicolumn{1}{l|}{\cellcolor[HTML]{FFFFFF}{\color[HTML]{000000} 0.52±0.09}} \\ \hline
\rowcolor[HTML]{EFEFEF} 
\multicolumn{1}{l|}{\cellcolor[HTML]{EFEFEF}{\color[HTML]{000000} NMI}} & \multicolumn{1}{l|}{\cellcolor[HTML]{EFEFEF}{\color[HTML]{000000} 0.11±0.08}} & \multicolumn{1}{l|}{\cellcolor[HTML]{EFEFEF}{\color[HTML]{000000} 0.08±0.07}} & \multicolumn{1}{l|}{\cellcolor[HTML]{EFEFEF}{\color[HTML]{000000} 0.05±0.07}} & \multicolumn{1}{l|}{\cellcolor[HTML]{EFEFEF}{\color[HTML]{000000} 0.07±0.08}} \\ \hline
\rowcolor[HTML]{FFFFFF} 
\multicolumn{1}{c}{\cellcolor[HTML]{FFFFFF}\textbf{}}                   & \multicolumn{4}{c}{\cellcolor[HTML]{FFFFFF}\textbf{GMM}}                                                                                                                                                                                                                                                                     \\
\rowcolor[HTML]{EFEFEF} 
\multicolumn{1}{l|}{\cellcolor[HTML]{FFFFFF}}                           & \multicolumn{1}{l|}{\cellcolor[HTML]{EFEFEF}Full}                             & \multicolumn{1}{l|}{\cellcolor[HTML]{EFEFEF}Spherical}                        & \multicolumn{1}{l|}{\cellcolor[HTML]{EFEFEF}Diagonal}                         & \multicolumn{1}{l|}{\cellcolor[HTML]{EFEFEF}Tied}                             \\ \hline
\rowcolor[HTML]{FFFFFF} 
\multicolumn{1}{l|}{\cellcolor[HTML]{FFFFFF}Sil}                        & \multicolumn{1}{l|}{\cellcolor[HTML]{FFFFFF}0.09±0.06}                        & \multicolumn{1}{l|}{\cellcolor[HTML]{FFFFFF}0.17±0.04}                        & \multicolumn{1}{l|}{\cellcolor[HTML]{FFFFFF}0.15±0.05}                        & \multicolumn{1}{l|}{\cellcolor[HTML]{FFFFFF}0.37±0.27}                        \\ \hline
\rowcolor[HTML]{EFEFEF} 
\multicolumn{1}{l|}{\cellcolor[HTML]{EFEFEF}MM}                         & \multicolumn{1}{l|}{\cellcolor[HTML]{EFEFEF}0.39±0.05}                        & \multicolumn{1}{l|}{\cellcolor[HTML]{EFEFEF}0.42±0.05}                        & \multicolumn{1}{l|}{\cellcolor[HTML]{EFEFEF}0.43±0.05}                        & \multicolumn{1}{l|}{\cellcolor[HTML]{EFEFEF}0.45±0.07}                        \\ \hline
\rowcolor[HTML]{FFFFFF} 
\multicolumn{1}{l|}{\cellcolor[HTML]{FFFFFF}FLK}                        & \multicolumn{1}{l|}{\cellcolor[HTML]{FFFFFF}0.39±0.05}                        & \multicolumn{1}{l|}{\cellcolor[HTML]{FFFFFF}0.40±0.05}                        & \multicolumn{1}{l|}{\cellcolor[HTML]{FFFFFF}0.41±0.04}                        & \multicolumn{1}{l|}{\cellcolor[HTML]{FFFFFF}0.47±0.14}                        \\ \hline
\rowcolor[HTML]{EFEFEF} 
\multicolumn{1}{l|}{\cellcolor[HTML]{EFEFEF}NMI}                        & \multicolumn{1}{l|}{\cellcolor[HTML]{EFEFEF}0.07±0.06}                        & \multicolumn{1}{l|}{\cellcolor[HTML]{EFEFEF}0.11±0.06}                        & \multicolumn{1}{l|}{\cellcolor[HTML]{EFEFEF}0.14±0.06}                        & \multicolumn{1}{l|}{\cellcolor[HTML]{EFEFEF}0.07±0.0}                         \\ \hline
\rowcolor[HTML]{FFFFFF} 
                                                                        & \multicolumn{4}{c}{\cellcolor[HTML]{FFFFFF}\textbf{Spectral}}                                                                                                                                                                                                                                                                 \\  
\rowcolor[HTML]{EFEFEF} 
\multicolumn{1}{l|}{\cellcolor[HTML]{FFFFFF}}                           & \multicolumn{1}{l|}{\cellcolor[HTML]{EFEFEF}40}                               & \multicolumn{1}{l|}{\cellcolor[HTML]{EFEFEF}60}                               & \multicolumn{1}{l|}{\cellcolor[HTML]{EFEFEF}80}                               & \multicolumn{1}{l|}{\cellcolor[HTML]{EFEFEF}100}                              \\ \hline
\rowcolor[HTML]{FFFFFF} 
\multicolumn{1}{l|}{\cellcolor[HTML]{FFFFFF}Sil}                        & \multicolumn{1}{l|}{\cellcolor[HTML]{FFFFFF}0.11±0.09}                        & \multicolumn{1}{l|}{\cellcolor[HTML]{FFFFFF}0.12±0.09}                        & \multicolumn{1}{l|}{\cellcolor[HTML]{FFFFFF}0.12±0.07}                        & \multicolumn{1}{l|}{\cellcolor[HTML]{FFFFFF}0.12±0.07}                        \\ \hline
\rowcolor[HTML]{EFEFEF} 
\multicolumn{1}{l|}{\cellcolor[HTML]{EFEFEF}MM}                         & \multicolumn{1}{l|}{\cellcolor[HTML]{EFEFEF}0.42±0.05}                        & \multicolumn{1}{l|}{\cellcolor[HTML]{EFEFEF}0.42±0.06}                        & \multicolumn{1}{l|}{\cellcolor[HTML]{EFEFEF}0.42±0.06}                        & \multicolumn{1}{l|}{\cellcolor[HTML]{EFEFEF}0.42±0.06}                        \\ \hline
\rowcolor[HTML]{FFFFFF} 
\multicolumn{1}{l|}{\cellcolor[HTML]{FFFFFF}FLK}                        & \multicolumn{1}{l|}{\cellcolor[HTML]{FFFFFF}0.42±0.06}                        & \multicolumn{1}{l|}{\cellcolor[HTML]{FFFFFF}0.42±0.06}                        & \multicolumn{1}{l|}{\cellcolor[HTML]{FFFFFF}0.42±0.06}                        & \multicolumn{1}{l|}{\cellcolor[HTML]{FFFFFF}0.42±0.06}                        \\ \hline
\rowcolor[HTML]{EFEFEF} 
\multicolumn{1}{l|}{\cellcolor[HTML]{EFEFEF}NMI}                        & \multicolumn{1}{l|}{\cellcolor[HTML]{EFEFEF}0.14±0.07}                        & \multicolumn{1}{l|}{\cellcolor[HTML]{EFEFEF}0.14±0.08}                        & \multicolumn{1}{l|}{\cellcolor[HTML]{EFEFEF}0.13±0.08}                        & \multicolumn{1}{l|}{\cellcolor[HTML]{EFEFEF}0.14±0.08}                        \\ \hline
\end{tabular}
\label{tab:clusterhyper}
\end{sidewaystable}

\subsection{Clustering algorithm selection}
To select the clustering method with the highest performance, we applied the GMM and KMeans algorithms to 2D and 3D postural information and appearance information on both datasets. The results are show in Tables \ref{tab:clusterhyper2} and \ref{tab:Bclusterhyper1}.

\begin{sidewaystable}[]
\caption{Clustering postural information using sampling techniques for Home (H) and HomeLab(HL) dataset using KMeans with L 1/2 distance function and GMM with tied covariance shape}
\begin{tabular}{cccccccccc}
\rowcolor[HTML]{FFFFFF} 
\multicolumn{2}{l}{\cellcolor[HTML]{FFFFFF}}                                                             & \multicolumn{4}{c}{\cellcolor[HTML]{FFFFFF}KMeans}                                         & \multicolumn{4}{c}{\cellcolor[HTML]{FFFFFF}GMM} \\ \cmidrule{3-10} 
\rowcolor[HTML]{FFFFFF} 
\multicolumn{2}{l}{\cellcolor[HTML]{FFFFFF}}                                                            & Sil       & MM        & FLK       & \multicolumn{1}{l}{\cellcolor[HTML]{FFFFFF}NMI}       & Sil        & MM         & FLK       & NMI       \\ \cmidrule{2-10} 
\rowcolor[HTML]{EFEFEF} 
\cellcolor[HTML]{FFFFFF}                          & \multicolumn{1}{l}{\cellcolor[HTML]{EFEFEF}2D - HL} & 0.50±0.29 & 0.50±0.07 & 0.52±0.07 & \multicolumn{1}{l}{\cellcolor[HTML]{EFEFEF}0.07±0.06} & 0.43±0.22  & 0.49±0.07  & 0.52±0.07 & 0.06±0.05 \\
\rowcolor[HTML]{FFFFFF} 
& \multicolumn{1}{l}{\cellcolor[HTML]{FFFFFF}3D - HL} & 0.82±0.28           &   0.55±0.09 & 0.61±0.09&  0.01 ±0.01 &\multicolumn{1}{l}{\cellcolor[HTML]{FFFFFF} 0.88±0.14} &   0.56±0.08&  0.64±0.06&   0.01±0.01\\ 
\rowcolor[HTML]{EFEFEF} 
\cellcolor[HTML]{FFFFFF}                          & \multicolumn{1}{l}{\cellcolor[HTML]{EFEFEF}2D - H} & 0.58±0.31 & 0.50±0.16 & 0.57±0.14 & \multicolumn{1}{l}{\cellcolor[HTML]{EFEFEF}0.05±0.07} & 0.44±0.30  & 0.48±0.14  & 0.53±0.13 & 0.05±0.06 \\
\rowcolor[HTML]{FFFFFF}                           & \multicolumn{1}{l}{\cellcolor[HTML]{FFFFFF}3D - H} & 0.87±0.19&0.61±0.11 &0.67±0.10& 0.01±0.01& \multicolumn{1}{l}{\cellcolor[HTML]{FFFFFF}  0.89±0.08}           & 0.62±0.11            & 0.68±0.08            &  0.01±0.02          \\ \cmidrule{2-10} 
\end{tabular}
\label{tab:clusterhyper2}
\end{sidewaystable}

\begin{sidewaystable}[]
\caption{Clustering appearance information using sampling techniques for Home(H) and HomeLab(HL) dataset using KMeans with L 1/2 distance function and GMM with tied covariance shape. }
\centering
\begin{tabular}{clllllllll}
\rowcolor[HTML]{FFFFFF} 
\multicolumn{1}{l}{\cellcolor[HTML]{FFFFFF}}      &                                                   & \multicolumn{4}{c}{\cellcolor[HTML]{FFFFFF}KMeans}                                             & \multicolumn{4}{c}{\cellcolor[HTML]{FFFFFF}GMM}                                                                                                   \\
\rowcolor[HTML]{FFFFFF} 
\multicolumn{1}{l}{\cellcolor[HTML]{FFFFFF}}      &                                                   & Sil        & MM         & FLK        & \multicolumn{1}{l}{\cellcolor[HTML]{FFFFFF}NMI}        & Sil                                & MM                                 & FLK                                & NMI                                \\ \cmidrule{2-10} 
\rowcolor[HTML]{EFEFEF} 
\cellcolor[HTML]{FFFFFF}                          & \multicolumn{1}{l}{\cellcolor[HTML]{EFEFEF}Max} & 0.27±0.06 & 0.48±0.09 & 0.45±0.09 & \multicolumn{1}{l}{\cellcolor[HTML]{EFEFEF}0.36±0.12} & 0.10±0.11                         & 0.45±0.07                         & 0.44±0.09                         & 0.38±0.12                          \\
\rowcolor[HTML]{FFFFFF} 
\multirow{-2}{*}{\cellcolor[HTML]{FFFFFF}\rotatebox[origin=c]{90}{HL}} & \multicolumn{1}{l}{\cellcolor[HTML]{FFFFFF}Mean}  & 0.34±0.02 & 0.48±0.07 & 0.46±0.05 & \multicolumn{1}{l}{\cellcolor[HTML]{FFFFFF}0.38±0.10} & 0.23±0.05                         & 0.46±0.06                         & 0.44±0.05                         & 0.26±0.11                         \\ 
\rowcolor[HTML]{EFEFEF} 
\cellcolor[HTML]{FFFFFF}                          & \multicolumn{1}{l}{\cellcolor[HTML]{EFEFEF}Max} & 0.09±0.04 & 0.46±0.07 & 0.45±0.07 & \multicolumn{1}{l}{\cellcolor[HTML]{EFEFEF}0.24±0.13} & \cellcolor[HTML]{EFEFEF}0.09±0.09 & \cellcolor[HTML]{EFEFEF}0.41±0.11 & \cellcolor[HTML]{EFEFEF}0.42±0.09 & \cellcolor[HTML]{EFEFEF}0.27±0.14 \\
\rowcolor[HTML]{FFFFFF} 
\multirow{-2}{*}{\cellcolor[HTML]{FFFFFF}\rotatebox[origin=c]{90}{H}}    & \multicolumn{1}{l}{\cellcolor[HTML]{FFFFFF}Mean}  & 0.33±0.13 & 0.46±0.10 & 0.45±0.09 & \multicolumn{1}{l}{\cellcolor[HTML]{FFFFFF}0.25±0.14} & 0.20±0.08                         & 0.44±0.08                          & 0.44±0.07                         & 0.25±0.12                         \\ \cmidrule{2-10} 
\end{tabular}
\label{tab:Bclusterhyper1}
\end{sidewaystable}

\subsection{2D pose estimation feature space construction}
Multiple feature representations using postural information from 2D poses were examined, as shown in the Table \ref{tab:clusterhyper21}. Certain grasps may take longer to manipulate objects compared to others, leading the clustering algorithm to form clusters biased towards the more time-consuming grasps and resulting in poor performance. To mitigate this issue, we employed a "uniform sampling technique," collecting a fixed number of data frames per task, up to 16 frames per task, regardless of the task duration.  

\begin{sidewaystable}[]
\caption{Clustering postural information using 2D poses for Home and HomeLab dataset using KMeans with L 1/2 distance function and GMM with tied covariance shape. W: thumb weighing. S: sampling method. C: including confidence.}
\begin{tabular}{llllllllll}
\rowcolor[HTML]{FFFFFF} 
\multicolumn{2}{l}{\cellcolor[HTML]{FFFFFF}}                   & \multicolumn{4}{c}{\cellcolor[HTML]{FFFFFF}KMeans}                                                                                                                                                   & \multicolumn{4}{c}{\cellcolor[HTML]{FFFFFF}GMM}                                                                                                                                                      \\ \cmidrule{3-10} 
\rowcolor[HTML]{FFFFFF} 
\multicolumn{2}{l}{\multirow{-2}{*}{\cellcolor[HTML]{FFFFFF}}} & \multicolumn{1}{c}{\cellcolor[HTML]{FFFFFF}Sil} & \multicolumn{1}{c}{\cellcolor[HTML]{FFFFFF}MM} & \multicolumn{1}{c}{\cellcolor[HTML]{FFFFFF}FLK} & \multicolumn{1}{c}{\cellcolor[HTML]{FFFFFF}NMI} & \multicolumn{1}{c}{\cellcolor[HTML]{FFFFFF}Sil} & \multicolumn{1}{c}{\cellcolor[HTML]{FFFFFF}MM} & \multicolumn{1}{c}{\cellcolor[HTML]{FFFFFF}FLK} & \multicolumn{1}{c}{\cellcolor[HTML]{FFFFFF}NMI} \\ \hline
\rowcolor[HTML]{EFEFEF} 
\cellcolor[HTML]{FFFFFF}                             & 2D       & 0.56±0.29                                       & 0.47±0.11                                      & 0.53±0.10                                       & 0.05±0.07                                        & 0.37±0.27                                       & 0.45±0.07                                      & 0.47±0.14                                       & 0.07±0.00                                        \\
\rowcolor[HTML]{FFFFFF} 
\cellcolor[HTML]{FFFFFF}                             & 2D W     & 0.58±0.28                                       & 0.47±0.11                                      & 0.37±0.23                                       & 0.09±0.09                                        & 0.37±0.23                                       & 0.49±0.09                                      & 0.49±0.09                                       & 0.08±0.10                                        \\
\rowcolor[HTML]{EFEFEF} 
\cellcolor[HTML]{FFFFFF}                             & 2D S     & 0.50±0.29                                       & 0.50±0.07                                      & 0.52±0.07                                       & 0.07±0.06                                        & 0.43±0.22                                       & 0.49±0.07                                      & 0.52±0.07                                       & 0.06±0.05                                        \\
\rowcolor[HTML]{FFFFFF} 
\cellcolor[HTML]{FFFFFF}                             & 2D C     & 0.35±0.23                                       & 0.45±0.08                                      & 0.46±0.09                                       & 0.10±0.11                                        & 0.36±0.31                                       & 0.48±0.09                                      & 0.49±0.09                                       & 0.11±0.08                                        \\
\rowcolor[HTML]{EFEFEF} 
\cellcolor[HTML]{FFFFFF}                             & 2D CW    & 0.27±0.20                                       & 0.44±0.06                                      & 0.44±0.06                                       & 0.12±0.07                                        & 0.36±0.26                                       & 0.46±0.10                                      & 0.50±0.10                                       & 0.15±0.14                                        \\
\rowcolor[HTML]{FFFFFF} 
\multirow{-6}{*}{\cellcolor[HTML]{FFFFFF}\rotatebox[origin=c]{90}{HomeLab}}    & 2D CWS   & 0.28±0.14                                       & 0.46±0.07                                      & 0.46±0.07                                       & 0.09±0.04                                        & 0.30±0.19                                       & 0.49±0.07                                      & 0.50±0.06                                       & 0.10±0.10                                        \\ \hline
\rowcolor[HTML]{EFEFEF} 
\cellcolor[HTML]{FFFFFF}                             & 2D       & 0.71±0.32                                       & 0.48±0.13                                      & 0.58±0.09                                       & 0.07±0.11                                        & 0.33±0.31                                       & 0.48±0.14                                      & 0.52±0.12                                       & 0.05±0.08                                        \\
\rowcolor[HTML]{FFFFFF} 
\cellcolor[HTML]{FFFFFF}                             & 2D W     & 0.80±0.30                                       & 0.49±0.12                                      & 0.58±0.08                                       & 0.02±0.06                                        & 0.22±0.22                                       & 0.49±0.13                                      & 0.51±0.12                                       & 0.05±0.10                                        \\
\rowcolor[HTML]{EFEFEF} 
\cellcolor[HTML]{FFFFFF}                             & 2D S     & 0.58±0.31                                       & 0.50±0.16                                      & 0.57±0.14                                       & 0.05±0.07                                        & 0.44±0.30                                       & 0.48±0.14                                      & 0.53±0.13                                       & 0.05±0.06                                        \\
\rowcolor[HTML]{FFFFFF} 
\cellcolor[HTML]{FFFFFF}                             & 2D C     & 0.31±0.18                                       & 0.51±0.09                                      & 0.51±0.08                                       & 0.10±0.10                                        & 0.32+0.25                                       & 0.49±0.11                                      & 0.53±0.08                                       & 0.07+0.11                                        \\
\rowcolor[HTML]{EFEFEF} 
\cellcolor[HTML]{FFFFFF}                             & 2D CW    & 0.46±0.30                                       & 0.51±0.12                                      & 0.54±0.11                                       & 0.10±0.11                                        & 0.41±0.33                                       & 0.49±0.11                                      & 0.54±0.11                                       & 0.05±0.10                                        \\
\rowcolor[HTML]{FFFFFF} 
\multirow{-6}{*}{\cellcolor[HTML]{FFFFFF}\rotatebox[origin=c]{90}{Home}}       & 2D CWS   & 0.30±0.22                                       & 0.45±0.07                                      & 0.47±0.09                                       & 0.05±0.05                                        & 0.29±0.22                                       & 0.50±0.15                                      & 0.52±0.13                                       & 0.05±0.06                                        \\ \hline
\end{tabular}
\label{tab:clusterhyper21}
\end{sidewaystable}

\section{Feature selection from appearance information using BYOL}\label{secA2}
The features extracted from Faster R-CNN may encode aspects of hand shape, hand colour, object geometry, and background information, but may not fully reflect meaningful semantic information related to hand grasps. Discriminative representation learning methods, such as contrastive methods, have recently attracted much attention. The key idea behind a contrasting algorithm is to increase the distance between the representation of different augmentations of different images while reducing the distance between different views from the same image. In other words, the majority of contrastive methods require positive and negative pairs. However, in practice, collecting many positive and negative instances involves some level of supervision \cite{khosla2020supervised}. Taking a different approach, BYOL \cite{grill2020bootstrap} and the SimSiam network have recently shown a tremendous capacity to learn representation without using negative pairs. Experimental reports on BYOL have suggested that batch normalization may implicitly introduce negative instances, avoiding model collapse \cite{tian2020understanding, grill2020bootstrap}. 

\begin{figure}[h!]
    \centering
    \includegraphics[width=\columnwidth]{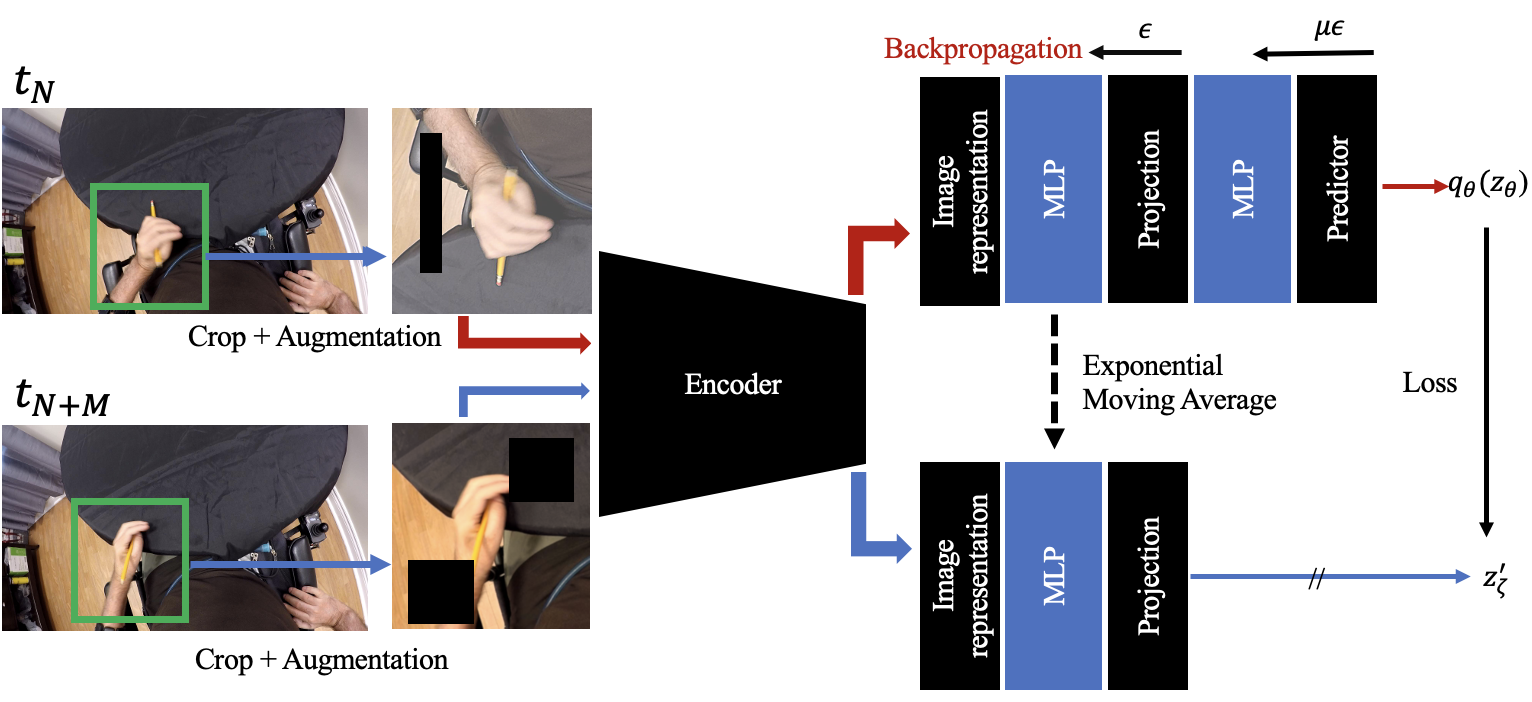}
        \caption{The proposed method is motivated by the BYOL algorithm using prediction and projection head. }
      \label{fig:method}
\end{figure}

The BYOL algorithm consists of a slow (target) and fast (online) network in which the prediction layer in the fast network predicts the target representation \cite{grill2020bootstrap}. The projection and prediction layers are multilayer perceptron networks with two hidden layers: a batch normalization layer and a ReLU activation function. Employing two views of the input images using a nonlinear augmentation algorithm, the feature space is computed using an encoder network. The projection layers in both target and slow networks map the features to another feature space. The predictor is trained to map the features from the online network to another feature space in which the distance between two representations from the target and the online network becomes minimized. In other words, the projector and predictor network allow us to estimate the representations invariant to some semantically meaningful augmentation. The proposed architecture is depicted in Figure \ref{fig:method}. 

Using the HOI method, the portion of the HomeLab dataset in which hands and objects are interacting was selected. The total number of raw images was 379,812 frames. The input images were augmented using the RandAugment method \cite{cubuk2020randaugment} by randomly selecting 2-4 augmentation methods described in table \ref{tab:augmentation}. We also augmented the image through time by assuming that the hand grasp during a stable grasp is constant. We estimated the interaction time using the HOI model \cite{shan2020understanding} and defined the first and last 10\% of the interaction as transitory grasps and the remaining 80\% as stable grasps. We randomly selected 256 pairs of images from the augmented images for each raw image. BYOL was reimplemented in Tensorflow. The BYOL algorithm presented here differs slightly from the original architecture, as fixed backbone weights were employed during the training in this paper. Grid search was performed on learning rate, number of neurons on fully connected layers, exponential moving average (EMA) coefficient, and learning rate coefficient for prediction head \(\mu\). The model was optimized using Layer-wise Adaptive Moments Based (LAMB) optimizer with the learning of 0.001 and eight GPUs, each with a mini-batch size of 512 using the mirror strategy in Tensorflow. Table \ref{tab:grid search} shows the grid for hyperparameters. After training the model with many augmented images, the projection or prediction layer was used for feature selection.

\begin{table}[]
    \centering
    \caption{ Augmentation methods}
    \begin{tabular}{l p{5cm} p{5cm}}
          Augmentation & Details & Magnitude\\
          \midrule
        Geometric:&Flipping right to left & NA\\
                  & Flipping up to down & NA\\
                  & Crop size& Normal(0.2, 0.15)*Length Image\\
                  & Random rotation & -50:+50\\
                  &Cutout&Normal(1,30)*Length Image
        \\
        \hline 
        \\
        Color space:& Random brightness & 0.5:1\\
                    & Random Hue & 0:0.16\\
                    & Random contrast  & 0.85 : 1.15\\
                    & Random saturation & 1: 1.15\\
                    & Random GaussianBlur&(10*10)\\
        \bottomrule
    \end{tabular}
    \label{tab:augmentation}
\end{table}

\begin{table}[]
    \centering
    \caption{ Candidate hyperparameters for BYOL algorithm}
    \begin{tabular}{p{5cm} p{4cm}}
        
          Hyperparameter & Candidate\\
          \midrule
         Representation layer& (Projection, Prediction)\\
         Condense layer& (Average, Max)\\
         Initial learning rate&(0.01, 0.001, 0.0001)\\
         Number of neurons ($\eta$) &(2048, 1024, 512)\\
         EMA coefficient ($\alpha$) & (0.950, 0.90)\\
         Learning rate coefficient ($\mu$)&(1, 5, 10, 15, 20)\\
        \bottomrule
    \end{tabular}
    \label{tab:grid search}
\end{table}

Next, we examined the subspace clustering with hyperparameters, described in detail in Table \ref{tab:grid search}. We identified the ones that yielded the best performance using IEI and EEI across all individuals. For each participant, the four highest examined hyperparameters with clustering performance results for both the Home and HomeLab datasets were selected to identify the best hyperparameter. The results showed that in 83\% of these cases, the models utilized a global mean layer, and in 56\% of instances, a prediction layer was employed. We found that augmenting the model with temporal information improved performance in 62\%  of cases. Moreover, setting the learning coefficient to 5 proved to be the most effective clustering method in 50\% of the best models. We also observed that configuring the number of neurons to 1024 achieved the highest performance in 88\% of the best models.  The best model performance for each clustering method for both datasets is shown in Table \ref{tab:Bclusterhyper2}.

\begin{table}[!htb]
\caption{Clustering appearance information using sampling techniques for Home and HomeLab dataset using KMeans with L 1/2 distance function and GMM with tied covariance shape using BYOL method}
\centering
\begin{tabular}{lcccccccc}
\rowcolor[HTML]{FFFFFF} 
\cellcolor[HTML]{FFFFFF}                             & \multicolumn{4}{c}{\cellcolor[HTML]{FFFFFF}K-Means}                    & \multicolumn{4}{c}{\cellcolor[HTML]{FFFFFF}GMM} \\ \cmidrule{2-9} 
\rowcolor[HTML]{FFFFFF} 
\multirow{-2}{*}{\cellcolor[HTML]{FFFFFF}}           & Sil  & MM   & FLK  & \multicolumn{1}{c|}{\cellcolor[HTML]{FFFFFF}NMI}  & Sil        & MM         & FLK       & NMI       \\ \hline
\rowcolor[HTML]{EFEFEF} 
\multicolumn{1}{l}{\cellcolor[HTML]{FFFFFF}HomeLab} & 0.36 & 0.48 & 0.45 & \multicolumn{1}{c|}{\cellcolor[HTML]{EFEFEF}0.25} & 0.27       & 0.48       & 0.46      & 0.32      \\ 
\rowcolor[HTML]{FFFFFF} 
\multicolumn{1}{l}{\cellcolor[HTML]{FFFFFF}Home}    & 0.17 & 0.46 & 0.46 & \multicolumn{1}{c|}{\cellcolor[HTML]{FFFFFF}0.22} & 0.21       & 0.45       & 0.43      & 0.24      \\ 
\end{tabular}
\label{tab:Bclusterhyper2}
\end{table}

\end{appendices}

\bibliography{sn-bibliography}
\end{document}